\documentclass{llncs}
\usepackage{makeidx}  
\usepackage{url}      
\usepackage{subfigure}
\usepackage{amsfonts}
\usepackage{amsmath}
\usepackage{cancel}
\usepackage{graphicx}
\usepackage{color}
\usepackage{listings}
\usepackage{hyperref}
\hypersetup{
    colorlinks=true,
    linkcolor=blue,
    filecolor=magenta,      
    urlcolor=blue,
}
 
\usepackage{tikz}
\usepackage{ifthen}
\usepackage{xcolor}
\usepackage{multirow}
\usetikzlibrary{shadows.blur}
\usepackage{booktabs}
\usepackage{xspace}

\newlength{\forkmeoffset}
\definecolor{forkmebg}{HTML}{CC0000}
\definecolor{forkmefg}{HTML}{EEEEEE}

\usepackage{graphicx}
\graphicspath{{pics/}{figs/}}

\begin{document}

\frontmatter          
\pagestyle{headings}  
\mainmatter              

\title{Correlation via synthesis: end-to-end nodule image generation and radiogenomic map learning based on generative adversarial network}
\titlerunning{Correlation via synthesis: end-to-end radiogenomic map learning based on generative adversarial network}  %

\author{
Ziyue Xu, Xiaosong Wang, Hoo-Chang Shin, Dong Yang, Holger Roth, Fausto Milletari, Ling Zhang, Daguang Xu
}

\institute{
Nvidia}
\authorrunning{}

\maketitle              

\begin{abstract}
Radiogenomic map linking image features and gene expression profiles is useful for noninvasively identifying molecular properties of a particular type of disease. Conventionally, such map is produced in three separate steps: 1) gene-clustering to “metagenes”, 2) image feature extraction, and 3) statistical correlation between metagenes and image features. Each step is independently performed and relies on arbitrary measurements. In this work, we investigate the potential of an end-to-end method fusing gene data with image features to generate synthetic image and learn radiogenomic map simultaneously. To achieve this goal, we develop a generative adversarial network (GAN) conditioned on both background images and gene expression profiles, synthesizing the corresponding image. Image and gene features are fused at different scales to ensure the realism and quality of the synthesized image. We tested our method on non–small cell lung cancer (NSCLC) dataset. Results demonstrate that the proposed method produces realistic synthetic images, and provides a promising way to find gene-image relationship in a holistic end-to-end manner.
\end{abstract}

\section{Introduction}
\label{sec:intro}


The integration of genome and imaging findings has emerged as a promising direction in clinical research, often being referred to as “radiogenomics”. A few studies has been performed to examine its potential in several diseases with different imaging techniques, including magnetic resonance (MR) in brain tumor \cite{Diehn2008}, and computed tomography (CT) in non–small cell lung cancer (NSCLC) \cite{Zhou2018}.

Despite the differences in disease and imaging modality, most current studies shared a common methodology in radiogenomics map generation. It is often designed following three independent steps: 1) image feature extraction, either computationally derived \cite{Gevaert2012}, or manually annotated \cite{Zhou2018}; 2) metagene clustering from the gene expression data on the basis of coexpression; and 3) statistical analysis to identify associations between image features and metagenes.

Although it is shown to be a viable way for radiogenomics, it may potentially have some limitations. First, image features are either existing hand-crafted sets or manually defined semantic judgements. The former may not be an optimal representation for the candidate data, and the later can in addition suffer from inter- and intra- observer variability. Second, the metagene clustering is based on statistical correlation analysis, which depends on the specific model being used, and may miss some correlation during model application. Third and most importantly, the image features and genetic characteristics are treated independently without knowledge of each other, and they can only be correlated in the last step. Hence, sub-optimal image representation, and model-dependent genome clustering may lead to weak correlations that can have limited power in reflecting reality. To out best knowledge, there is no prior work treating all three steps in a holistic, end-to-end manner.      
 
With recent development of deep learning, image features can be learnt from data and be optimized for a specific task \cite{Shin2016}. Comparing with hand-crafted or semantic features, such learnt feature presents higher accuracy and robustness in several tasks. Meanwhile, generative adversarial network (GAN) has enabled computer vision researchers to artificially generate realistic images from noise, reference images, and word embeddings \cite{Zhang2017}. A few successes have also been achieved in medical domain \cite{Jin2018}. GAN features the capability of fusing information from different sources to generate the output.  

The purpose of this work is to investigate the potential of a multi-conditional GAN designed for holistically analyze gene expression data and medical images. By utilizing them for new sample generation, both image features and gene embeddings can be learnt directly from data in an end-to-end manner. As a proof-of-concept study, we applied our strategy to a public NSCLC dataset with gene expression profiles from RNA sequencing. NSCLC is a common type of lung cancer and leading cause of mortality, and it is known that both imaging and gene expression play important role in its management. 

The major contributions of this work are 1) we formulate image-gene correlation by solving a multi-conditional GAN, 2) we design a new GAN architecture and fusion blocks to combine image (as background) and gene (as object and ``style'') information, 3) smooth object/background fusion is modeled within network rather than formulated as ``inpainting'', and 4) we demonstrate that a discriminative radiogemonic map can be learnt via this synthesis strategy.

\section{Method}
\label{sec:method}

In this work, we approach radiogemonic map by formulating it as an image synthesis task. Existing method for CT lung nodule simulation are mostly modeled as ``inpainting'' based on conditional GAN with no \cite{Jin2018} \cite{Liu2018} or limited \cite{Yang2018} ability in combining other information beyond the surrounding image. There are two major challenges in applying such network for radiogemonic purpose: 1) inpainting removes part of the image making space for the simulated nodule, therefore the regional information is lost and hence it is difficult to ensure spatial continuity and to avoid artificial looking around the boundary of erased region; and 2) there is no direct mechanism to introduce non-image information to the network, which is required for genomic data. 

To address these issues, inspired by computer vision works for natural image synthesis \cite{Park2018} \cite{Karras2018}, we design our network as a multi-conditional GAN with style specification. Foreground/background fusion is modeled within network, while image/gene coding are both utilized for synthesis. Fig.\ref{fig:noduleGAN} illustrates an overview of our method. Below, we outline the GAN architecture, information fusion design, and training strategy for learning the representation and generating lung nodules from both imaging and genomic data.
 
\subsection{GAN architecture}
\begin{figure*}[htbp!]
\vspace{-10pt}
\centering
\includegraphics[width=0.8\textwidth]{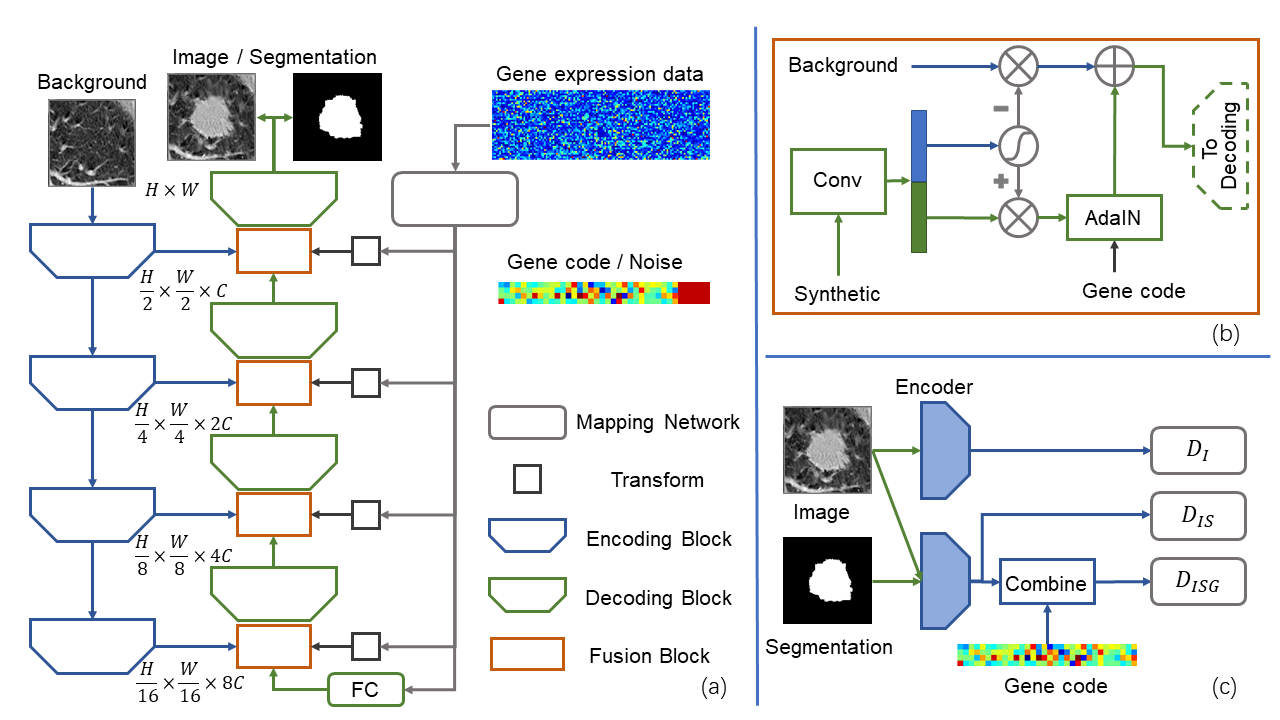} 
\setlength{\belowcaptionskip}{-15pt}
\caption{Proposed multi-conditional GAN for radiogenomic map learning and nodule synthesis. (a) Generator utilizes both background image and gene code to synthesize image together with nodule segmentation. (b) Fusion block at each resolution layer helps to fuse the information from background with that from previous layer and gene code. (c) With image, segmentation, and gene code, discriminator distinguishes three types of real/fake scenarios.}
\label{fig:noduleGAN}
\end{figure*}

Fig.\ref{fig:noduleGAN} (a) illustrates the structure of the proposed generator. From background image and gene expression data, it generates a synthetic image with a nodule characterized by the genomic data, and situated within the background image. Meanwhile, it also produces a binary segmentation mask of the generated nodule. Structure-wise, it consists of three parts: encoding of the background image (left), encoding of gene expression data (right), and information fusion for synthetic image/mask generation (center). 

The proposed GAN handles two major challenges: object/background separation and blending, and image/gene representation fusion.

For the first challenge, unlike previous inpainting-based methods, the proposed network does not remove any portion of the background image. Instead, it models object/background within the network via two strategies: 1) a fusion block at each resolution level to control the overlapping between generated object with reference background, and 2) an auxiliary output of segmentation mask to guild such separation. At each stage, it performs a ``soft'' blending of object/background information, and therefore ensures spatial continuity of the outcome synthetic image. Another strength of this method is that it produces segmentation mask together with the image, and is hence potentially helpful for other tasks such as detection and segmentation if being used as a data augmentation technique \cite{Jin2018}.    

For the second challenge, existing work in computer vision \cite{Park2018} used word embedding to produce a base image that can be combined with background at the bottleneck layer of a encoder-decoder network. One major difference between word embedding and gene representation is that word is much more closely related to image. Therefore, we choose to model the gene information as the abstract ``style'' of an image, and use style transfer techniques to guide the synthesis process \cite{Karras2018}. Specifically, high-dimensional gene expression data is encoded with a mapping network, this can be a few fully-connected (FC) layers \cite{Karras2018}, or more sophisticated conditioning augmentation block \cite{Zhang2017}\cite{Park2018}. Here, for better interpretability of gene encoding, we choose to use two FC layers to encode the raw gene data $g$ to a vector $\phi(g)$. $\phi(g)$ is further concatenated with a noise vector $n$ to be used as the base style map. Meanwhile, the background image is encoded with conventional image encoders consists of convolutional layers \cite{Park2018}. With image features and gene map, we use a series of fusion blocks to combine the information. The fusion block take image features from both background and previous step, together with gene ``style'' map to achieve: 1) proper blending of object/background, and 2) proper fusion of image/gene information. 


\subsection{Information fusion}
Fig.\ref{fig:noduleGAN} (b) described the proposed fusion block. At each resolution level, we have three information feed-ins: background image feature, gene map, and synthesis image feature from the previous layer. Since it contains information for both object and background, the synthetic features are first further encoded via two layers of convolution and batch normalization. During this process, the channel number is doubled. The resulting code is then split into two parts: the first half used as a weight map to control how much object/background information will be passed to further processing at this layer; while the other half will be used as object feature map. 

As shown in the figure, both object/background feature maps will be controlled by element-wise multiplication with the weight map ($+$) and its inverse ($-$). Map $+$ suppressed the background information, passing mainly nodule features to be normalized by gene code. This is because gene code has less to do with background and more to control nodule appearance. Meanwhile, map $-$ suppressed the information where nodule will be generated, reinforcing background information to align with input image. Gene code then controls the ``style'' of the synthetic nodule via an adaptive instance normalization (AdaIN) layer \cite{Karras2018}. Finally, the two are added together and fed to upsampling/decoding layer. 

Comparing with completely erasing part of the image as ``inpainting'', the weight map is a learnt probability, which retains the information necessary for smooth fusion of object and background. Comparing with word embedding synthesis \cite{Park2018}, our strategy strengthened the object/background separation because the gene map is to be applied mostly to the object region and has little impact over background. 

\begin{figure*}[htbp!]
\vspace{-10pt}
\centering
\includegraphics[width=0.8\textwidth]{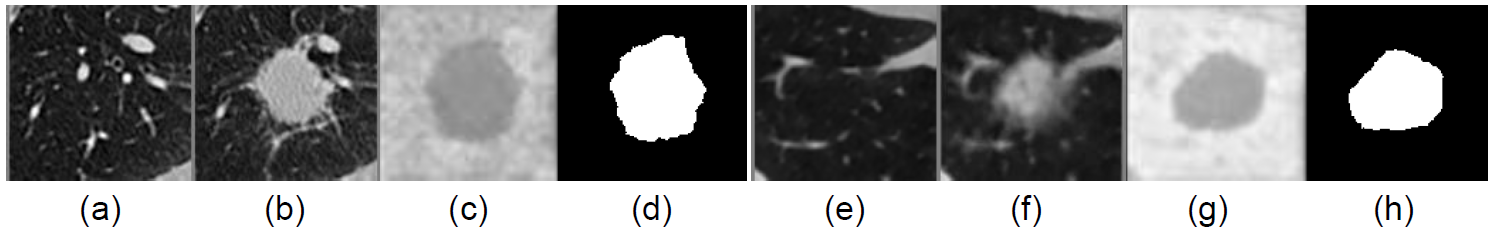}
\setlength{\belowcaptionskip}{-15pt}
\caption{Examples of proposed synthesis GAN: (a, e) background image, (b, f) synthesized nodule image, (c, g) background weight image, (d, h) segmentation mask. 
\label{fig:example}}
\end{figure*} 

Fig.\ref{fig:example} shows two examples (a-d) and (e-h) for the proposed GAN. (a) is the background image, (b) is the synthesis result, (c) is the background weight map, and (d) is the resulting mask. (e-h) is the same as (a-d) but for another case with ground-glass opacity. It can be observed that the original background image does not get change significantly with major structures preserved. At the same time, the synthesized nodule fused naturally with background image, even for a ground-glass case. It can also be seen that for two background images under different reconstruction, (e) being smoother than (a), the sharpness of the resulting nodule image also aligns well with the background image.

\subsection{Training strategy}
The proposed GAN encodes genomic features as a vector, and outputs both image and segmentation. Here, the discriminator is illustrated in Fig.\ref{fig:noduleGAN} (c), following the method in \cite{Park2018}. The input to the discriminator is a tuple of image-segmentation-gene code. Two encoders are utilized to encode: 1) image for discriminator $D_{I}$, and 2) image-segmentation pairs for discriminator $D_{IS}$. The second encoder's output is further combined with gene code $\phi(g)$ and further encoded via convolution, batch normalization, and leaky ReLU activation layers for discriminator ${D_{ISG}}$. Discriminators are trained with least squares loss functions \cite{Mao2017}. Given image $x$, matched gene code $g$, and matched segmentation mask $m$, tuples to be discriminated against it include cases containing mismatched gene code $\bar{g}$, mismatched segmentation mask $\bar{m}$, synthetic image $G_x$, and synthetic mask $G_m$. Let $p_d$ and $p_G$ denote the distributions of real and synthetic data, we have $x, g, m, \bar{g}, \bar{m} \sim p_d$ and $G_x, G_m \sim p_G$. With different combinations, we have
\begin{align*}
L_{D_{I}} &= \mathbb{E}[(D_{I}(x)-1)^2] + \mathbb{E}[D_{I}(G_x)^2] \\
L_{D_{IS}} &= \mathbb{E}[(D_{IS}(x,m)-1)^2] + \mathbb{E}[D_{IS}(x,\bar{m})^2] + \mathbb{E}[D_{IS}(G_x,G_m)^2] \\
L_{D_{ISG}} &= \mathbb{E}[(D_{ISG}(x,m,g)-1)^2] + \mathbb{E}[D_{ISG}(x,\bar{m},g)^2] \\ 
& + \mathbb{E}[D_{ISG}(x,m,\bar{g})^2] + \mathbb{E}[D_{ISG}(G_x,G_m,g)^2]
\end{align*}

For training the generator, the background reconstruction loss is added to guide the feature extraction of background image during synthesis. Let $G_{\bar{M}}$ be a morphological eroded version of segmentation mask $G_m$'s inverse (i.e. background region), $\odot$ denote element-wise multiplication, the $L1$ loss is computed over background between synthetic image $G_x$ and base image $x$:
\begin{align*}
L_{G} &= \mathbb{E}[(D_{I}(G_x)-1)^2] + \mathbb{E}[(D_{IS}(G_x,G_m)-1)^2] \\
& + \mathbb{E}[(D_{ISG}(G_x,G_m,g)-1)^2] + \lambda\mathbb{E}[\|G_x\odot G_{\bar{M}} - x\odot G_{\bar{M}}\|_1]
\end{align*}

\section{Experiments and Results}
\label{sec:experiments}
We evaluate the proposed method using the publicly available NSCLC dataset \cite{NSCLC}. This radiogenomic dataset is built upon a NSCLC cohort of 211 subjects. Together with CT images and  segmentation maps of the tumors, a subset of 130 subjects also have RNA sequencing data from surgically excised tumor tissue samples. After removing all gene with NaN values, we end up with a 5172-dimensional gene vector for each case. A $60\times60\times60 \ \textrm{mm}^3$ volume-of-interest (VOI) centered at each nodule is first cropped from the original image. Then 2D slices with nodule presence are extracted as training samples. In total we have 3736 training image slices. 

Background images are created as following: lung region is first segmented for each image, then the nodule regions are excluded from lung mask. Next, distance transform is computed for the resulting mask, and centers are selected at a random location 5 to 25 mm from the mask boundary. Around each center, a $60\times60\times60 \ \textrm{mm}^3$ VOI is cropped and 20 random slices are extracted from each VOI.

\begin{figure*}[htbp!]
\vspace{-5pt}
\centering
\includegraphics[width=0.95\textwidth]{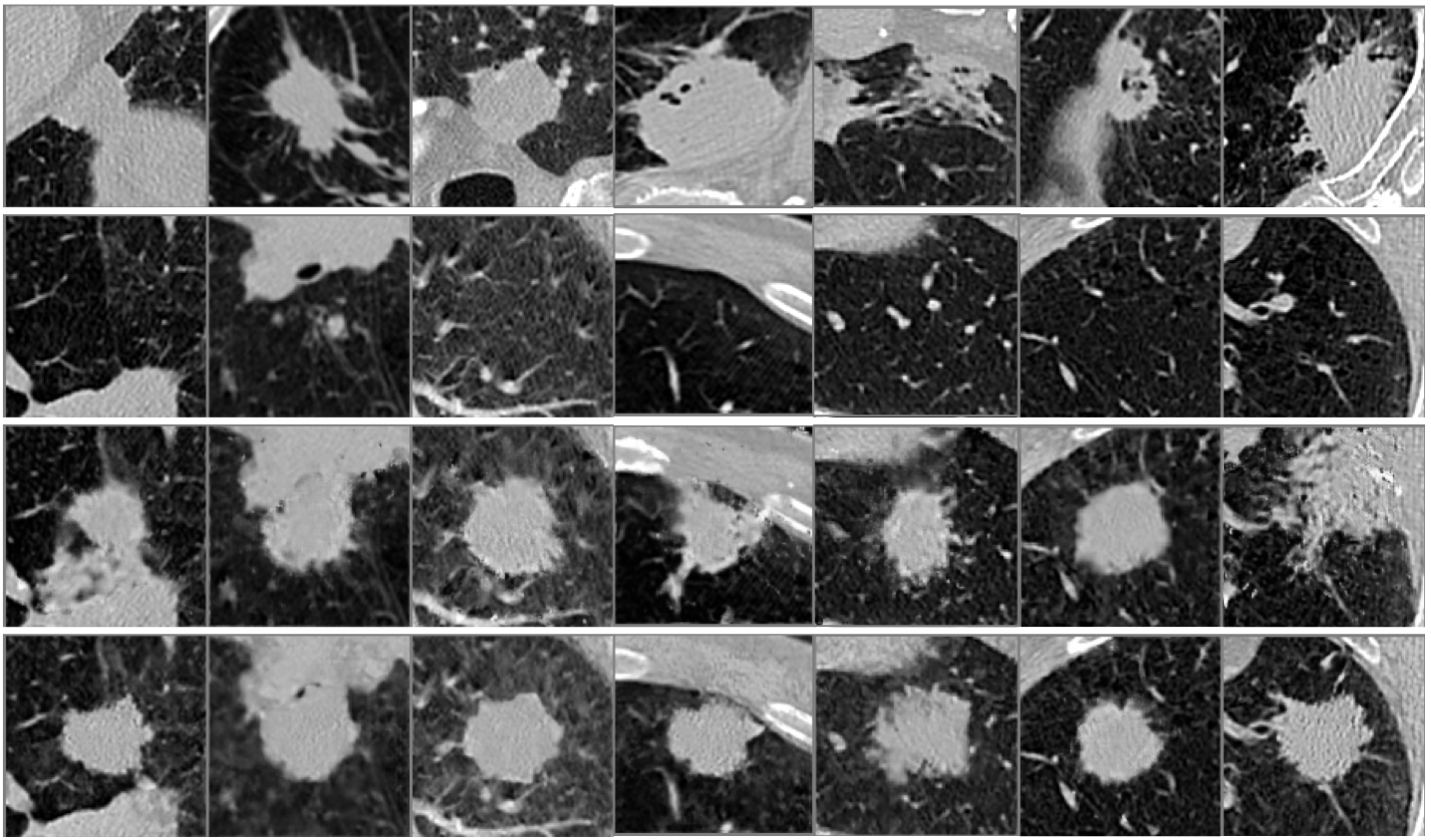} 
\vspace{-5pt}
\setlength{\belowcaptionskip}{-15pt}
\caption{Result of nodule synthesis, first row: training image, whose genomic information is used to synthesize each column; second row: background image; third row: synthetic image generated by baseline method \cite{Park2018}; last tow: synthetic image generated by the proposed method.}  
\label{fig:result}
\end{figure*}

Our proposed method has two goals: 1) realistic and controlled generation of nodule images, and 2) radiogemonic map learning that links gene information to image features. Since there is no prior work achieving these goals, we compare the proposed method against baseline method \cite{Park2018} (2D multi-conditional natural image synthesis).

Fig. \ref{fig:result} shows the performance of image synthesis with multi-conditional GAN. First row include 7 training images. The algorithm uses the gene information from each of them, together with background images from second row, to synthesize image with nodules. Third row is the results from baseline method, while the last row is from the proposed method. As shown in the image, the proposed method generates more realistic images than baseline, and the resulting synthetic images have similar features as the reference training samples.

\begin{figure*}[htbp!]
\vspace{-5pt}
\centering
\includegraphics[width=\textwidth]{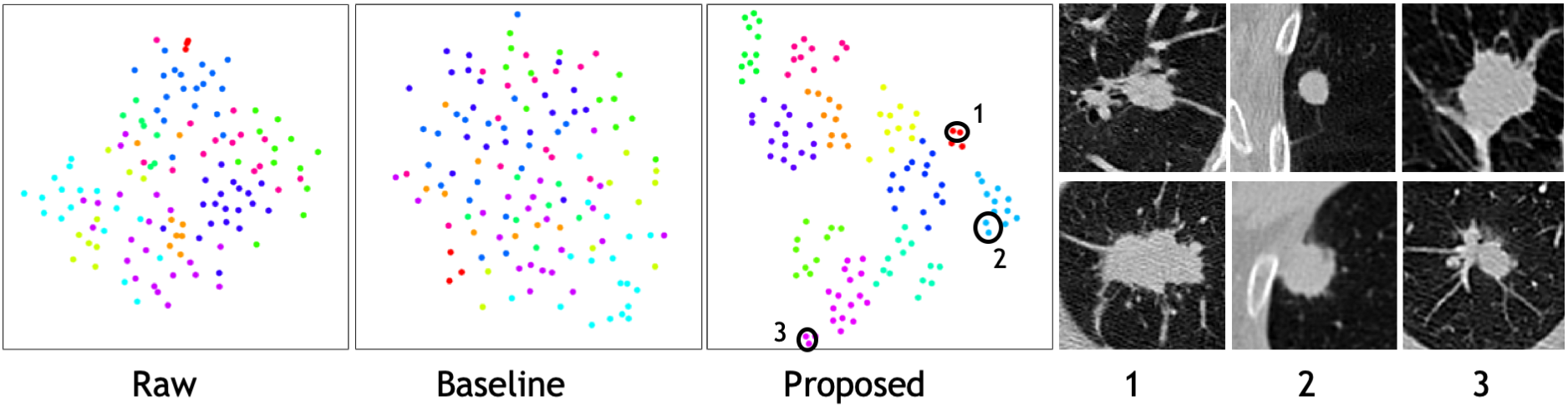} 
\vspace{-5pt}
\setlength{\belowcaptionskip}{-10pt}
\caption{Distribution of gene coding illustrated by 2D t-SNE map \cite{Maaten2008} : raw gene (5172-D) and gene code produced by baseline method (128-D) does not show obvious separation, while gene code produced by the proposed method (128-D) showed feasibility for clustering. Three groups of samples are drawn from clusters formed according to distance, and their corresponding image are shown. 
\label{fig:cluster}}
\end{figure*}

Fig. \ref{fig:cluster} shows the radiogenomic correlation of the trained network. We use t-SNE \cite{Maaten2008} to generate a 2D representation of genomic data. As shown, no clusters can be reasonably formed from the raw gene vector, or the one produced by baseline method. With the proposed method, the resulting gene code can be separated to clusters. By examining their corresponding image, we can observe some general correlation between gene cluster and the image features such as nodule shape and boundary smoothness.

\section{Conclusion}
\label{sec:conclusions}
We use a multi-conditional GAN, coupled with a new structure of style control and fusion, to effectively generate realistic nodules whose appearance is controlled by its genomic features. Without erasing any portion of condition image, our method is superior over state-of-the-art method in object realism and background fusion. An end-to-end mechanism is achieved to holistically model and correlate various features. As such, our approach can provide not only an effective and controllable means to generate diverse nodules, but also a discriminative radiogemonic map linking genomic and image features. Currently, this work is proof of concept given that the data size is limited. More data and experiments will be needed to further validate this approach for radiogemonic study.

\bibliographystyle{splncs03}
\bibliography{bibliography}

\begin{thebibliography}{10}
\providecommand{\url}[1]{\texttt{#1}}
\providecommand{\urlprefix}{URL }

\bibitem{NSCLC}
Bakr, S., Gevaert, O., Echegaray, S., Ayers, K., Zhou, M., Shafiq, M., Zheng,
  H., Zhang, W., Leung, A., Kadoch, M., Shrager, J., Quon, A., Rubin, D.,
  Plevritis, S., Napel, S.: {Data for NSCLC Radiogenomics Collection} (2017),
  the Cancer Imaging Archive

\bibitem{Diehn2008}
Diehn, M., Nardini, C., Wang, D.S., McGovern, S., Jayaraman, M., Liang, Y.,
  Aldape, K., Cha, S., Kuo, M.D.: {Identification of noninvasive imaging
  surrogates for brain tumor gene-expression modules}. Proceedings of the
  National Academy of Sciences  105(13),  5213--5218 (2008)

\bibitem{Gevaert2012}
Gevaert, O., Xu, J., Hoang, C.D., Leung, A.N., Xu, Y., Quon, A., Rubin, D.L.,
  Napel, S., Plevritis, S.K.: {Non–Small Cell Lung Cancer: Identifying
  Prognostic Imaging Biomarkers by Leveraging Public Gene Expression Microarray
  Data—Methods and Preliminary Results}. Radiology  264(2),  387--396 (2012)

\bibitem{Jin2018}
Jin, D., Xu, Z., Tang, Y., Harrison, A.P., Mollura, D.J.: {CT-Realistic Lung
  Nodule Simulation from 3D Conditional Generative Adversarial Networks for
  Robust Lung Segmentation}. In: Medical Image Computing and Computer Assisted
  Intervention -- MICCAI 2018. pp. 732--740. Springer International Publishing,
  Cham (2018)

\bibitem{Karras2018}
Karras, T., Laine, S., Aila, T.: {A Style-Based Generator Architecture for
  Generative Adversarial Networks}. CoRR  abs/1812.04948 (2018)

\bibitem{Liu2018}
Liu, S., Gibson, E., Grbic, S., Xu, Z., Setio, A.A.A., Yang, J., Georgescu, B.,
  Comaniciu, D.: {Decompose to manipulate: Manipulable Object Synthesis in 3D
  Medical Images with Structured Image Decomposition}. CoRR  abs/1812.01737
  (2018)

\bibitem{Maaten2008}
van~der Maaten, L., Hinton, G.: {Visualizing High-Dimensional Data Using
  t-SNE}. Journal of Machine Learning Research  9,  2579--2605 (Nov 2008)

\bibitem{Mao2017}
{Mao}, X., {Li}, Q., {Xie}, H., {Lau}, R.Y.K., {Wang}, Z., {Smolley}, S.P.:
  {Least Squares Generative Adversarial Networks}. In: 2017 IEEE International
  Conference on Computer Vision (ICCV). pp. 2813--2821 (Oct 2017)

\bibitem{Park2018}
Park, H., Yoo, Y., Kwak, N.: {MC-GAN: Multi-conditional Generative Adversarial
  Network for Image Synthesis}. In: The British MachineVision Conference (BMVC)
  (2018)

\bibitem{Shin2016}
{Shin}, H., {Roth}, H.R., {Gao}, M., {Lu}, L., {Xu}, Z., {Nogues}, I., {Yao},
  J., {Mollura}, D., {Summers}, R.M.: {Deep Convolutional Neural Networks for
  Computer-Aided Detection: CNN Architectures, Dataset Characteristics and
  Transfer Learning}. IEEE Transactions on Medical Imaging  35(5),  1285--1298
  (May 2016)

\bibitem{Yang2018}
Yang, J., Liu, S., Grbic, S., Setio, A.A.A., Xu, Z., Gibson, E., Chabin, G.,
  Georgescu, B., Laine, A.F., Comaniciu, D.: {Class-Aware Adversarial Lung
  Nodule Synthesis in {CT} Images}. CoRR  abs/1812.11204 (2018)

\bibitem{Zhang2017}
Zhang, H., Xu, T., Li, H., Zhang, S., Wang, X., Huang, X., Metaxas, D.N.:
  {StackGAN: Text to Photo-Realistic Image Synthesis With Stacked Generative
  Adversarial Networks}. In: The IEEE International Conference on Computer
  Vision (ICCV) (Oct 2017)

\bibitem{Zhou2018}
Zhou, M., Leung, A., Echegaray, S., Gentles, A., Shrager, J.B., Jensen, K.C.,
  Berry, G.J., Plevritis, S.K., Rubin, D.L., Napel, S., Gevaert, O.:
  {Non–Small Cell Lung Cancer Radiogenomics Map Identifies Relationships
  between Molecular and Imaging Phenotypes with Prognostic Implications}.
  Radiology  286(1),  307--315 (2018)

\end{thebibliography}

\end{document}